\documentclass[sn-basic]{sn-jnl}

\jyear{2022}%

\theoremstyle{thmstyleone}%
\theoremstyle{thmstyletwo}%

\theoremstyle{thmstylethree}%

\begin{document}

\title[A Large-Scale Analysis of Persian Tweets Regarding Covid-19 Vaccination]{A Large-Scale Analysis of Persian Tweets Regarding Covid-19 Vaccination}


\author[1]{\fnm{Taha} \sur{ShabaniMirzaei}}\email{taha.shabani@ut.ac.ir}

\author[1]{\fnm{Houmaan} \sur{Chamani}}\email{houmaan.chamani@ut.ac.ir}

\author[1]{\fnm{Amirhossein} \sur{Abaskohi}}\email{amir.abaskohi@ut.ac.ir}

\author[1]{\fnm{Zhivar} \sur{Sourati Hassan Zadeh}}\email{zh.sourati@ut.ac.ir}

\author*[1]{\fnm{Behnam} \sur{Bahrak}}\email{bahrak@ut.ac.ir}

\affil*[1]{\orgdiv{Department of Electrical and Computer Engineering}, \orgname{University of Tehran}, \orgaddress{\city{Tehran}, \country{Iran}}}


\abstract{
The Covid-19 pandemic had an enormous effect on our lives, especially on people's interactions. By introducing Covid-19 vaccines, both positive and negative opinions were raised over the subject of taking vaccines or not. In this paper, using data gathered from Twitter, including tweets and user profiles, we offer a comprehensive analysis of public opinion in Iran about the Coronavirus vaccines. For this purpose, we applied a search query technique combined with a topic modeling approach to extract vaccine-related tweets. We utilized transformer-based models to classify the content of the tweets and extract themes revolving around vaccination. We also conducted an emotion analysis to evaluate the public happiness and anger around this topic. Our results demonstrate that Covid-19 vaccination has attracted considerable attention from different angles, such as governmental issues, safety or hesitancy, and side effects. Moreover, Coronavirus-relevant phenomena like public vaccination and the rate of infection deeply impacted public emotional status and users' interactions.
}




\maketitle

\section{Introduction}
\label{sec:intorduction}

The first officially known outbreak of the Covid-19 was initiated in Wuhan, China, at the end of 2019 \citep{site:who2021}. According to the rapid dissemination of Coronavirus and the number of lost lives from this infection, the Covid-19 pandemic has massively impacted our daily lives, interactions, behaviors, and routines. Although upcoming breakouts are potential and the future of the Covid-19 pandemic is uncertain \citep{Bonnevie2021}, currently, there are several vaccines which act as controlling measures for the disease outbreak.

As mentioned by \cite{ThanhLe2020}, controlling factors other than the quality of the vaccines, such as public support and trust towards authorities, are essential to ensure the efficiency of vaccination programs. However, these types of treatments, particularly those offered as an emergency response to a rapidly spreading pandemic, are sometimes looked upon with reservation and reluctance \citep{TROIANO2021245}. Therefore, with respect to the overall aim of global immunization and prevention of social consequences of the pandemic, there exists great potential and need for studies that analyze both supportive and critical viewpoints related to mass vaccination of the population. Understanding critical viewpoints and their rationales is helpful to convince a wider proportion of society into getting vaccinated and increasing the success rate of such programs worldwide.

Nowadays, social media platforms play a significant role in our lives. People communicate, express their feelings and passions, and inform or get informed about the latest news via these platforms. Investigating social media can shed light on measuring people's attitudes toward any discussed topic and recognizing how their opinions evolve over time. In recent years, Twitter has been a key source of information dissemination as one of the most powerful social networks. Each user on Twitter can broadcast a message that may contain any desired content, as long as he/she abides by the platform's safety, privacy, and authenticity rules\footnote{\url{https://help.twitter.com/en/rules-and-policies/twitter-rules}}.

Despite the fact that content on Twitter is publicly accessible, conducting research on tweets requires a detailed plan for acquiring and analyzing relevant data. This paper presents a practical approach for mining and classification of Persian tweets and users regarding Coronavirus vaccination, leading to a detailed analysis of public supportive and critical attitudes on vaccination in Iran. Furthermore, our research centers on the Persian language, a resource-constrained linguistic system that has received considerably less scrutiny in the realm of social studies when compared to English. In addition, this research provides insights into the relationship between different events and social media reactions to them. 
The contribution of this paper can be summarized as follows:
\begin{itemize}
    \item We describe a topic modeling approach combined with a keyword-based method for extracting Persian tweets related to vaccination.
    \item We apply transformer-based machine learning techniques for tweet classification.
    \item We conduct an emotion analysis using the labelled dataset for happiness and anger emotions in Persian words.
    \item We quantify different supportive and critical vaccination themes extracted from tweets.
    \item We investigate users' connections before and after the initiation of vaccination.
\end{itemize}

The remainder of this paper is organized as follows: Section \ref{sec:related-work} gives a brief synopsis of the previous related works. Afterwards, Section \ref{sec:data} explains how Persian tweets relevant to Covid-19 have been collected. In Section \ref{sec:methods}, we present the preprocessing methodologies as well as our approaches for obtaining tweets related to vaccination, and introduce a strategy to classify the tweets into three classes: negative, positive, and neutral. Techniques used for emotion analysis and further evaluations, such as extracting vaccine themes and user study, are also explored in this section. Section \ref{sec:results} analyzes classified tweets and extracted themes. Furthermore, multiple pieces of analysis about the Covid-19 timeline, user groups and influential users, and overall emotion analysis results are included in this section. Then, Section \ref{sec:conclusion} concludes the paper and outlines future research directions. Finally, Section \ref{sec:discussion} addresses the study's limitations, discusses the practical implications of our results for various audiences, and highlights how our findings extend prior research.


\section{Related Work}
\label{sec:related-work}

Considering the diversity, richness, and availability of Twitter data, several pieces of research are conducted utilizing tweets to analyze the impact of Covid-19 on societies and social media platforms \citep{kwok2021tweet, lyu2022social, cascini2022social}. According to Covid-19 Data Explorer \footnote{\url{https://ourworldindata.org/explorers/coronavirus-data-explorer}}, Iran was one of the first countries got infected by Covid-19; nevertheless, there are only a few analyses carried out to investigate Iranians' opinions toward Coronavirus and vaccination. \cite{HOSSEINI2020} has performed one of the early studies conducted to gauge responses to ongoing events by categorizing Persian tweets into different classes and demonstrating how the reactions evolved over time. Besides, \cite{SHOKROLLAHI2021} provides a Post-structuralist Discourse Analysis (PDA) of the Covid-19 phenomenon in Persian society using social network graphs to cluster and explore influencers. Moreover, sentiment analysis of Persian tweets related to Covid-19 has been conducted in this piece of research. Lastly, \cite{NEZHAD2022} presented a sentiment analysis approach to assess Persian community's position toward domestic and imported Coronavirus vaccines. 

Generally, topic detection can help structure an extensive data collection by grouping records into different classes. In order to achieve a reliable classification, many topic modeling techniques are available. \cite{LYE2021E24435} aims to identify the topics of tweets related to Covid-19, fetched with relevant keywords, using Latent Dirichlet Allocation (LDA) topic modeling developed by \cite{LDA2003993}. Similarly, \cite{WICKE2021} employs LDA to illustrate how the subjects linked with the pandemic growth change over time. On the other hand, we compared LDA with Gibbs Sampling for Dirichlet Multinomial Mixture (GSDMM) from \cite{GSDMM2014233} as the first-step in classifying Persian tweets. GSDMM is a modified LDA technique mainly used for short text topic modeling (STTM) tasks, assuming only one topic for each document rather than a probability distribution on all the potential topics from the original LDA. We have considered both LDA and GSDMM models and compared their results to extract the most relevant topics. 

One important factor for analyzing public opinions toward vaccination is to explore trends and reactions during the pandemic. According to the temporal evolution study of different emotional categories and influencing factors implemented in \cite{CHOPRA2021}, expressing doubt about vaccination attracts the highest health-related conversations in all the countries studied during the research. Furthermore, \cite{Thelwall_Kousha_Thelwall_2021} applies a manual content analysis on a small portion of vaccine-hesitant Coronavirus tweets in English to extract major themes discussed regarding hesitancy. Likewise, quantifications introduced in \cite{BONNEVIE202112} compare vaccine-critical posts on Twitter before and after the Covid-19 spread in the United States, which depicts a significant increase in vaccine disapproval, especially in areas related to health authorities, vaccine ingredients, and research trials. Moreover, in \cite{BONNEVIE2020S326}, vaccine opposition themes are manually coded, and afterward, misinformation in each theme, as well as top influencers, are identified. The results show that prominent influencers appear to be well coordinated in misinformation dissemination. Apart from vaccine trends, another direction of our study is to classify vaccine-related tweets into three categories and discuss the evolution of each position (critical, supportive, and neutral) during the pandemic.

In addition to vaccination topics, there are pieces of research conducted on sentiment analysis of tweets with respect to the Covid-19 vaccination. One example is \cite{WICKE2021}, which performs sentiment analysis based on the Pattern library, which uses a dictionary of manually-tagged adjectives with values for sentiment polarity in tweets \cite{JMLR:v13}. Similarly, \cite{YOUSEFINAGHANI2021256} utilizes Valence Aware Dictionary and sEntiment Reasoner (VADER), a Python lexicon and rule-based sentiment analysis tool, to assign sentiment polarity to every tweet \cite{Hutto_Gilbert_2014}. Furthermore, in a recent study, \cite{NEZHAD2022} applied a deep learning model reinforced with a sarcasm detection approach to achieve high accuracy for Persian tweets.

Although several projects were carried out for vaccine themes identification and sentiment analysis, many plausible analyses in these areas received less attention, especially in Persian, which is a low-resource language. In previous studies, the main concentration has usually been on vaccine-opposition themes, while we explore themes and demonstrate how they develop throughout time using a grounded theory methodology devised by \cite{grounded2014}. Furthermore, we perform emotion analysis over different prominent vaccination opinions, i.e., positive, negative, and neutral, using our tagged Persian words emotion dataset.

As for the focus on studying users involved in Covid-19 related conversations, one of the first studies was carried out by \cite{BONNEVIE2020S326}. By analyzing “Top Authors” and user engagement, they found that vaccine opposition and misinformation does not come from a diverse distribution of users. Additionally, \cite{YOUSEFINAGHANI2021256} has classified Twitter users into three categories, namely pro-vaccine, anti-vaccine, and neutral and determined how each user belongs to each group. A similar study for the Turkish Twitter has been conducted by \cite{durmaz2022}. A key point in their work is that they identified anti-vaccine influencers both before and after the pandemic. As for the study at hand, we use a robust method to categorize each user into the positions mentioned above and study user interactions after and before the public vaccination in Iran.


\section{Data}
\label{sec:data}
\begin{figure}[ht] 
\centering
\includegraphics[scale=.45]{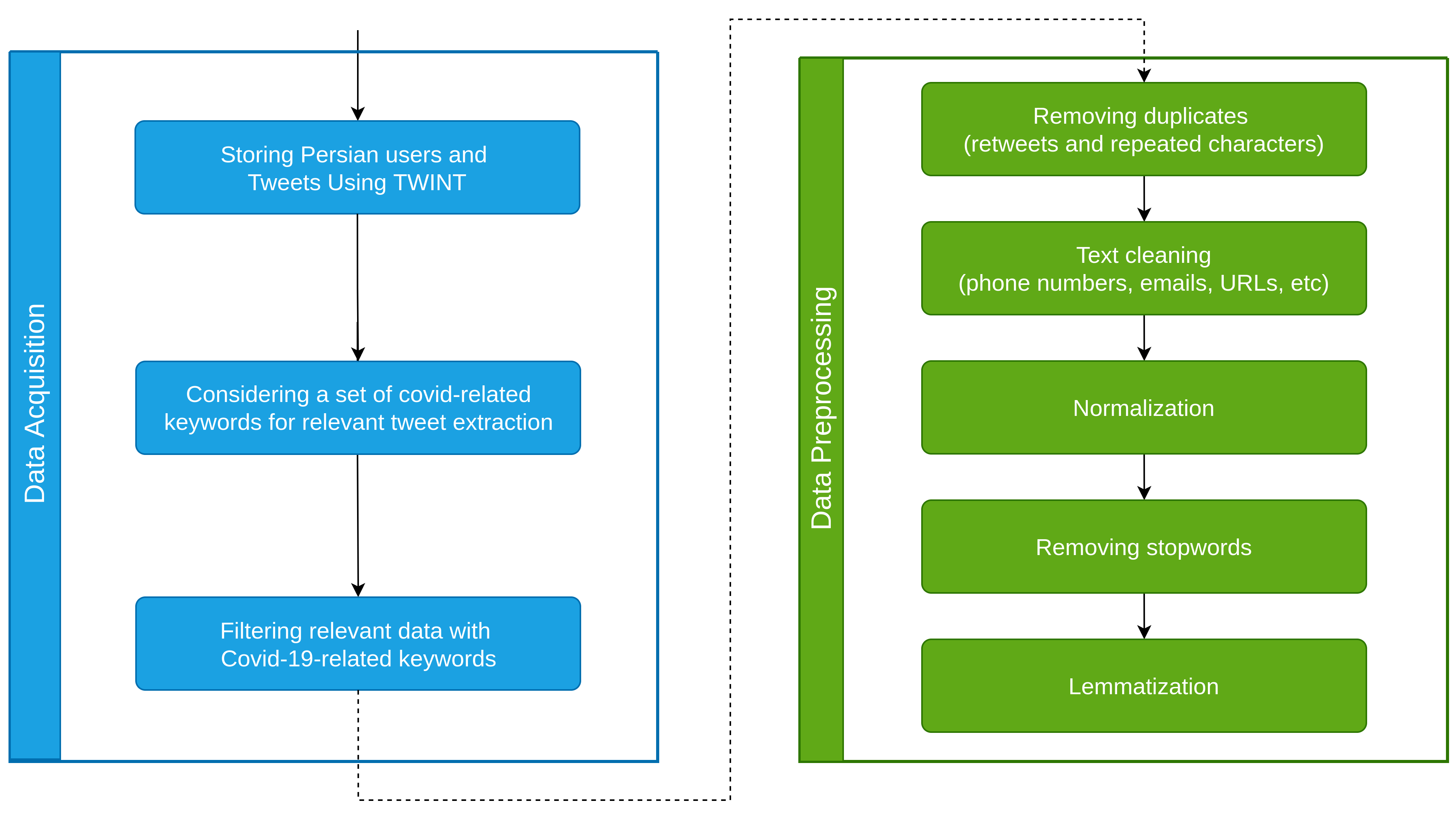}
\caption{Data Acquisition and Preprocessing} 
\label{fig:data_procedure} 
\end{figure}

As previously stated, this study aims to analyze Persian tweets about vaccination to provide insight into the public opinion toward Coronavirus vaccines in Iran. In order to fulfill this goal, we first need to collect relevant data for processing. The data acquisition and preprocessing procedures are fully explored in the following according to the workflow shown in Figure \ref{fig:data_procedure}.

\subsection{Data Acquisition}
To collect Persian tweets and their respective users, we have not just focused on our task at hand; instead, we have gathered a comprehensive dataset to be potentially utilized for further studies. This dataset contains 709,460,922 tweets and 6,661,480 active users from Jan. 2012 to Dec. 2021.


In this endeavor, we used Twitter Intelligence Tool (TWINT), which is an advanced Twitter scraping tool developed by \cite{site:TWINT}, allowing us to gather Twitter users' profiles and tweets. We modified TWINT so that we could extract users' and tweets' information for every hour and saved them in Elasticsearch. We chose Elasticsearch as our database and search engine because of its robustness and scalable architecture.

In the next step and in order to separate Covid-19-related tweets, we extracted tweets from Feb. 2020, when the first infected case in Iran was publicly announced, up until Dec. 2021, based on at least one of the following keywords in Persian: \textbf{Corona, Covid, vaccine, and quarantine}. More information on the number of tweets for each keyword is provided in Table \ref{tab:covid_tweets}.

\begin{table}[ht] 
\caption{Keywords and Related Tweets} 
\centering
\begin{tabular}{|c|c|}
\hline\hline
 Keywords & \  Tweet Count \\ [0.25ex] 
\hline
Corona & 2,673,287 \\
Covid & 72,987 \\
Vaccine & 958,250 \\
Quarantine & 398,142 \\
\hline
At least one keyword & 3,825,742 \\
\hline\hline  
\end{tabular} 
\label{tab:covid_tweets} 
\end{table}

Extracted information contains features for users and tweets. We have stored the list of mentioned users in a tweet and whether a tweet is a reply to another one, along with the count of users' interactions with the tweet. For example, we have only saved the number of likes per tweet, not the list of users who liked the tweet, since we were only interested in the quantity of this statistic. Similarly, the number of followers and followings has been gathered for each user, but the list of the followers or followings is not available. Table \ref{tab:tweet_features} and Table \ref{tab:user_features} describe further details regarding the main properties of tweets and users datasets, respectively.

\begin{table}[ht] 
\caption{Description of Tweet Features} 
\centering
    \begin{tabular}{|p{0.23\linewidth} | p{0.67\linewidth}|}
\hline\hline
 Feature Name & \  Description \\ [0.25ex] 
\hline
Tweet ID & Unique ID for every tweet \\
User ID & Unique ID employed by Twitter for the owner of the tweet \\
Conversation ID & ID employed by Twitter for the conversation \\
Retweet Count & Number of retweets \\
Reply Count & Number of replies \\
Like Count & Number of likes \\
Reply to & This field contains the User ID of the replied tweet if the current tweet is a reply to another tweet  \\
Mentions & List of mentioned user IDs \\
Created at & Creation time of the tweet \\
Source & Twitter Source (Android, iPhone, iPad, Web App) \\
Hashtags & List of hashtags in the tweet \\
URLs & List of URLs in the tweet \\
Tweet & Tweet content \\
\hline\hline  
\end{tabular} 
\label{tab:tweet_features} 
\end{table} 

\begin{table}[ht] 
\caption{Description of User Features} 
\centering
    \begin{tabular}{|p{0.23\linewidth} | p{0.67\linewidth}|}
\hline\hline
 Feature Name & \  Description \\ [0.25ex] 
\hline
ID & Unique ID employed by Twitter for every user \\
Username & The name that identifies the user \\
Bio & Biography of the user \\
Location & Location of the user \\
URL & Link in the user account \\
Joined Time & Time of account creation \\
Tweet Count & Number of tweets \\
Like Count & Number of total likes\\
Followers & Number of followers\\
Followings & Number of followings\\
Private & Whether user account is private\\
Verified & Whether user account is verified\\
\hline\hline  
\end{tabular} 
\label{tab:user_features} 
\end{table}

\subsection{Data Preprocessing}
\label{subsec:dp}
With inspiration from \cite{Philippines2021}, who performed a comprehensive data preprocessing over tweets in the Philippines, we have improved the procedure of data analytics by devising a pipeline for cleaning and preprocessing, consisting of multiple states. The scheme of preprocessing is shown in Figure \ref{fig:data_procedure}. Further details are aptly explained in the following:

\medbreak
\subsubsection{Removing Duplicates}

Afterward, we removed duplicate records with similar tweet content and one-word tweets because these tweets often do not imply any meaningful concepts. After this phase, 286,546 records were eliminated.

\medbreak
\subsubsection{Text Cleaning}
We used the Clean-Text library in Python \citep{site:CLEAN_TEXT} in addition to our customized techniques for data cleaning. Clean-Text was employed to provide a better text representation. We used this library to fix various Unicode errors and remove URLs, phone numbers, emails, and currency symbols. Moreover, we also removed HTML tags as well as meaningless characters and punctuation.

\medbreak
\subsubsection{Normalization}
In our text preprocessing workflow, normalization plays a vital role in ensuring the consistency and standardization of Persian text data. We employed the Hazm library \citep{site:HAZM}, a specialized tool tailored for processing Persian text, to carry out this essential task. The main objective of normalization is to unify various classes of terms within the text, simplifying their representation and enhancing text analysis. This process includes tasks such as removing diacritics, correcting spacing issues, and ensuring that the text adheres to standard conventions. By utilizing the Hazm Normalizer, we effectively achieved these normalization goals, resulting in more consistent and standardized Persian text data for our analysis.

\medbreak
\subsubsection{Removing Stopwords}
We defined a set of Persian stopwords to be removed from tweets using a combination of the Hazm stopwords dataset \citep{site:HAZM} and Persian stopwords defined in \cite{site:PERSIAN_STOPWORDS}. Afterward, we investigated every word in these two sets and removed those that might be relevant to Covid-19 and vaccination. Finally, we evaluated the top-appearing words in Covid-19-related tweets and checked whether they referred to any meaningful notion; if not, we appended them to our stopwords set.

\medbreak
\subsubsection{Lemmatization}
We also performed lemmatization on our Persian dataset using the Hazm lemmatizer to reduce inflections and variant forms to their base form. Considering that lemmatization can alter or even reverse the meaning of words, especially in transforming negative verbs into infinitives, we created two datasets, one with lemmatization (LEM) and the other without it (N-LEM), to compare the impact of lemmatization on topic modeling results and subsequent steps.

\section{Methods}
\label{sec:methods}

In order to figure out a way to filter tweets relevant to vaccination, we used a topic modeling approach combined with a keyword-based search. We also applied a transformer-based machine learning technique to classify vaccine-related tweets into three major groups (vaccine-critical, vaccine-supportive, and neutral).

\begin{figure}[ht] 
    \centering
    \includegraphics[scale=.5]{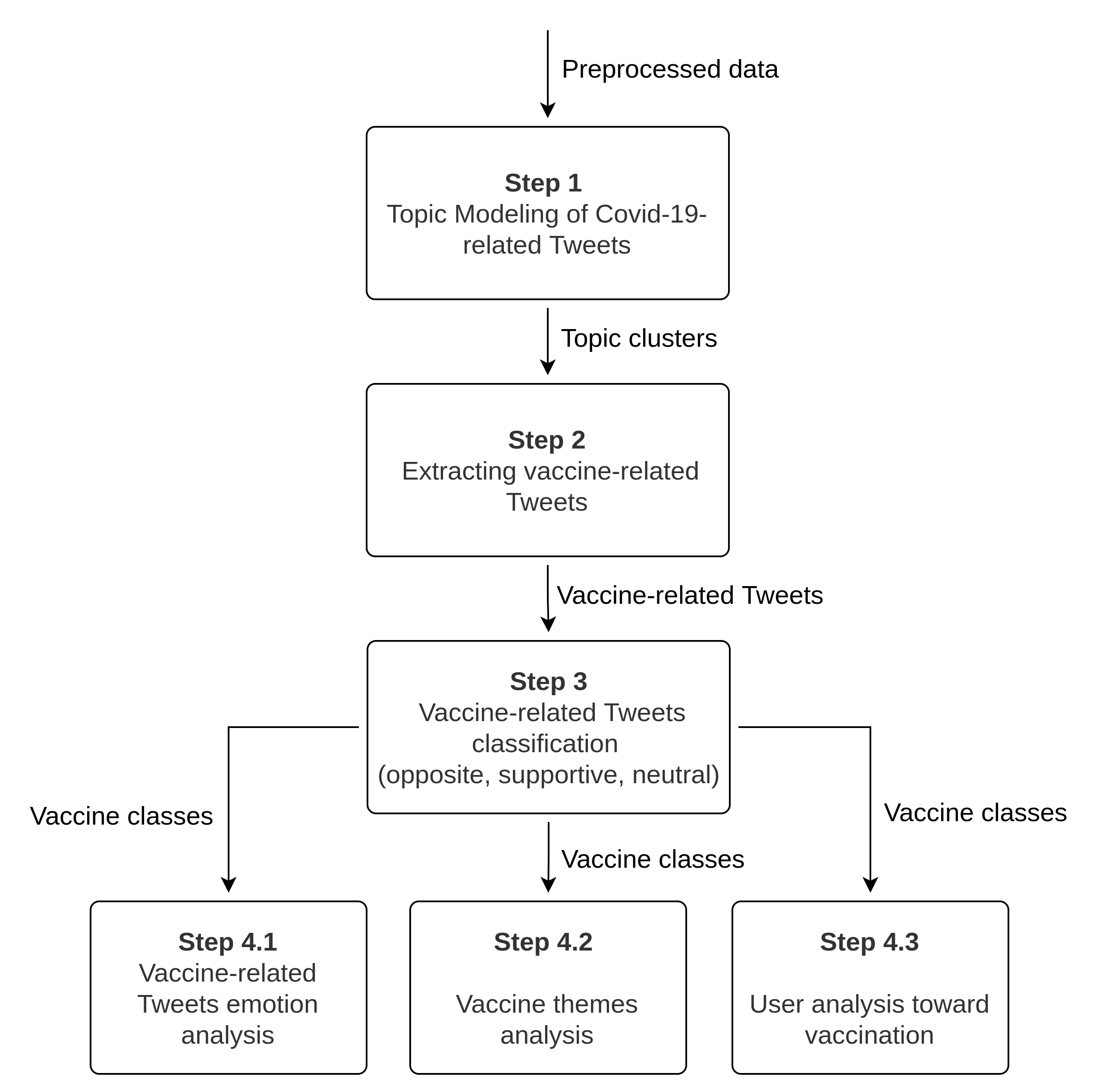}
    \caption{Workflow of Twitter Analysis toward Covid-19 Vaccination} 
    \label{fig:method_workflow} 
\end{figure}

Additional details about the exploited research methodology is shown in Figure \ref{fig:method_workflow}.

\subsection{Topic Modeling}

Topic modeling, also referred to as probabilistic clustering, is an approach to structure a large dataset and classify it into smaller, more interpretable, and spatially separated clusters. There are many topic modeling methodologies available, of which we chose LDA, which is an unsupervised machine learning algorithm and the most widely used technique, and GSDMM, an approach for short-text classification tasks. We applied these two topic modeling techniques to our dataset and compared their results to see which one performs better.
 
We used a combination of two criteria to assess the performance of our topic modeling algorithms:
\begin{enumerate}
    \item \textbf{Coherence measure ($C_v$) by \cite{CV2010100}:} topic coherence measures calculate the degree of semantic similarity between high-scoring terms in a topic to determine its score. These metrics aid in distinguishing between semantically and non-semantically interpretable issues.
    
    \item \textbf{Human judgment}: similar to what \cite{NIPS2009_f92586a2} has proposed, we carried out the word and topic intrusion tasks, focusing on the meaning of the words in subjects to examine topics and assess the interpretability of each group. These tasks involved human evaluators assessing the meaningfulness and coherence of topics based on linguistic and contextual cues. They quantitatively evaluated the semantic meaning of inferred topics using novel methods backed by large-scale user studies. Importantly, this approach reveals aspects of topic quality that traditional measures like held-out likelihood might miss. The research emphasizes that practitioners often assume that the latent space in topic models has semantic meaning but highlights the need for quantitative evaluation of this interpretability. Notably, it reveals that models excelling in held-out likelihood may yield less semantically meaningful topics, indicating a disconnect between conventional measures and human judgment. Our evaluation method incorporates these insights, prioritizing human judgment to assess the interpretability of topics generated by our models

\end{enumerate}

In order to achieve the most reasonable topic models, we evaluated several factors over a sample of 100,000 tweets. First, we compared the LEM dataset with N-LEM based on the coherence value over changes in multiple hyper-parameters and word representations. LEM dataset outperformed N-LEM on an average of 2.3\% in $C_v$ score over 25 executions. Because of the mentioned reason, we opted to use the LEM dataset for the rest of the topic modeling process.

Next, we compared Bag-of-Words (BoW) and Term Frequency/Inverse Document Frequency (TF-IDF) word representation techniques. For this purpose, we filtered out extreme tokens that appeared in fewer than 15 tweets or more than 50\% of all tweets and kept only the top 100,000 tokens for topic modeling execution. On an average of 20 executions, BoW results were 2.2\% better than TF-IDF.

Finally, we tuned LDA and GSDMM hyper-parameters to find the best results for each method. The parameters that gave the best results are described below.

LDA parameters:
\begin{itemize}
    \item $NT$: The number of themes to be retrieved from the training corpus.
    \item $NP$: It signifies the count of times the model passes through the training corpus during the training process.
    \item $\alpha$: A number for a symmetric prior over document-topic distribution.
    \item $CS$: Number of documents/tweets in each training chunk.
\end{itemize}

GSDMM parameters:
\begin{itemize}
    \item $NT_G$: The upper limit for the number of topics. 
    \item $NI$: It sets the maximum number of GSDMM algorithm iterations, with each iteration involving data point reevaluation and potential cluster reassignment through Gibbs sampling for model improvement.
    \item $\alpha_G$: A parameter ranging from 0 to 1, controlling records' affinity for a larger cluster.
    \item $\beta_G$: A parameter ranging from 0 to 1, controlling records' affinity for a more homogeneous cluster.
\end{itemize}

We evaluated results for $NT$ (and $NT_G$) between 5 and 10 and $NS$ (and  $NI$) between 6 and 12 for LDA and GSDMM models. Based on the $C_v$ coherence measures shown in Table \ref{tab:lda_gsdmm_coherences} and human judgments, LDA model outperforms GSDMM on our dataset.

\begin{table}[ht] 
\caption{Best Results Gained from Topic Modeling}
\centering
    \begin{tabular}{|c|c|c|}
\hline\hline
 Topic model & \  Number of Topics & \  Coherence (C\_v) \\ [0.25ex] 
\hline
LDA & 10 & 52.72\% \\
GSDMM & 9 & 42.46\% \\
\hline\hline  
\end{tabular} 
\label{tab:lda_gsdmm_coherences} 
\end{table}

After finding the best model for the tweets using the LDA technique, we manually labeled each group according to the concept perceived from each cluster. More information about these topics is provided in Table \ref{tab:lda_topic_details}.

\begin{table}[ht] 
\caption{Final Topics} 
\centering
    \begin{tabular}{|c|c|c|}
\hline\hline
 Topic description & \  Number of Tweets & \  \% of all tweets \\ [0.25ex] 
\hline
Religious and governmental & 257,314 & 7.27\% \\
Relatives and mourning & 370,644 & 10.47\% \\
Vaccination opinions & 527,294 & 14.90\% \\
Regional news & 293,378 & 8.29\% \\
Reports and statistics & 188,088 & 5.31\% \\
Symptoms & 501,034 & 14.16\% \\
Political and dissatisfaction & 161,774 & 4.57\% \\
quarantine and education & 456,551 & 12.90\% \\
Vaccination (news, reports) & 424,406 & 12.00\% \\
Political and financial & 358,713 & 10.13\% \\
\hline\hline  
\end{tabular} 
\label{tab:lda_topic_details} 
\end{table}

\subsection{Vaccine-related Tweets}

Keyword-based search is usually practical for providing a required subset; however, it only relies on the presence of a list of words. Consequently, there was a lack of implication and sentence meaning when utilizing keywords to provide data. In order to address this challenge and obtain the most relevant tweets related to Covid-19 vaccination, we developed a hybrid approach and merged the results obtained from the keyword-based technique with our topic modeling outcomes.

According to our topic modeling results, two groups were related to vaccination, i.e., vaccination opinions and vaccination news and reports. First, we extracted tweets with a high probability of belonging to one of these two clusters, defining a probability of more than or equal to 0.5 as high. Based on this criteria, 499,228 tweets were extracted from the dataset.

Then, we defined a series of vaccine-related keywords, whose translations to English are as follows: \textbf{vaccine, vaccination, Astra, AstraZeneca, Pfizer, Moderna, Sputnik, Covaxin, Sinopharm}. The rest of the Covid-19-related tweets were checked against these words. 538,212 tweets contained at least one of these keywords. Consequently, we stored 1,037,440 tweets related to vaccination for further studies.

\subsection{Vaccine-related Tweets Classification}

After providing vaccine-related tweets, our goal was to classify them into three major groups: vaccine-critical, neutral, and vaccine-supportive. To achieve this, we initially manually labeled 6000 tweets using the grounded theory approach. For the first 1000 items of the extracted dataset, the first two authors independently labeled the tweets into the three categories mentioned above. Then, the two labeled datasets were compared using Cohen's Kappa metric, indicating a consistency of 78 percent. Following a discussion about the tweets that didn't receive the same label, a consistency of 90 percent was achieved for the first 1000 labeled tweets. Subsequently, the remaining portion was divided into two datasets, each consisting of 2500 tweets and labeled by a single person. To ensure label accuracy, both annotators cross-checked each other's labels and engaged in discussions as necessary. Table \ref{tab:tweet-class-example} provides translated and paraphrased tweet examples to protect users' privacy and Table \ref{tab:polar_dist} presents the statistics of the labeled dataset.

\begin{table}[ht] 
\caption{Examples of Classified Vaccine-Related Tweets} 
\centering
    \renewcommand{\arraystretch}{1.5}
    \begin{tabular}{|p{0.70\linewidth} | p{0.21\linewidth}|}
\hline\hline
 Tweet & \  Class \\ [0.25ex] 
\hline
Dying within 10 days of getting the COVID-19 vaccine? It's so scary.  &  \\
Vaccine damages the brain and nerves \#no\_to\_vaccination. & Vaccine-critical \\
I get vaccinated with this sore throat and I'm gonna die.  &  \\
\hline
Some countries began the third vaccine dose while we're still waiting for the first. &  \\
Seminary students should take charge of making the vaccine. & Neutral \\
In my view, surviving COVID-19 is easier if you're in debt because people pray for your survival daily.  &  \\
\hline
When Lambda comes, everyone will be caught. However, it's worst for countries with less vaccinated people. &  \\
Finally, I got vaccinated :))) \#Sinofarm & Vaccine-supportive \\
It's very interesting that my friend says in the 21st century, I'm still not sure if I should get the vaccine or not?!! &  \\
\hline\hline  
\end{tabular} 
\label{tab:tweet-class-example} 
\end{table}

\begin{table}[ht] 
\caption{Polarity Distribution of Hand-Labeled Dataset} 
\centering
    \begin{tabular}{|c|c|c|}
\hline\hline
 Position & \  Count & \  Percentage \\ [0.25ex] 
\hline
Vaccine-Critical & 1735 & 28.9\% \\
Neutral & 2611  & 43.5\% \\
Vaccine-Support & 1654  & 27.5\% \\
\hline\hline  
\end{tabular} 
\label{tab:polar_dist} 
\end{table}

Subsequently, manually labeled data were used for vaccine opinion classification. We employed a combination of 5 different factors for data preprocessing, as shown in Table \ref{tab:dataset_extensions}. For text cleaning and removing stopwords, we considered three different criteria, namely extreme, moderate, and no filtering. The details of these three criteria are as follows:

\begin{itemize}
    \item Extreme: Applying all the methods mentioned in Section \ref{subsec:dp}.
    \item Moderate: Allowing the presence of vaccine-related words, for which we reduced the size of the stopwords set by 30\%. Also, for the text cleaning part, punctuations, numbers, and conversational forms were kept in tweets.
    \item No Filtering: Keeping tweet contents intact.
\end{itemize}

On the other hand, we considered only two possibilities for duplicate removal and lemmatization: whether to apply them or not. In this stage, we generated 36 different datasets from our original vaccine-related tweets.

\begin{table}[ht] 
\caption{Dataset Extension Criteria} 
\centering
    \begin{tabular}{|c|c|c|}
\hline\hline
 Criteria & \  States & \  \# of States \\ [0.25ex] 
\hline
Duplicate Removal & Keep / Remove & 2 \\
Text Cleaning & Extreme / Moderate / No Filter & 3 \\
Lemmatization & Apply / Ignore & 2 \\
Stopword Elimination & Extreme / Moderate / No Filter & 3 \\
\hline\hline  
\end{tabular} 
\label{tab:dataset_extensions} 
\end{table}

Finally, we applied transformer-based machine learning techniques to perform our classification of vaccine-related tweets. We fine-tuned and compared a series of these approaches with pre-trained models that used a Masked Language Modeling (MLM) objective to determine the best result. The strategies we employed are discussed in the following:

\medbreak
\subsubsection{Bidirectional Encoder Representations from Transformers (BERT)}

BERT, introduced in \cite{BERT2018}, applied bidirectional training of the transformer, a popular attention model, to language modeling. This method contrasted with previous endeavors, as it considered a text sequence from both left-to-right and right-to-left training modes. Initially, we used BERT-base and BERT-large models. Subsequently, we utilized ParsBERT from \cite{ParsBERT}, a monolingual language model based on Google’s BERT architecture, which had been pre-trained on large Persian corpora consisting of more than 3.9M documents, 73M sentences, and 1.3B words. Similar to previous models, we fine-tuned ParsBERT v3.0 and compared the results with BERT-base and BERT-large.

\medbreak
\subsubsection{Robustly Optimized BERT Pretraining Approach (RoBERTa)}

\cite{ROBERTA2019} trained BERT with more input data and epochs, resulting in RoBERTa, which demonstrated the effectiveness of these techniques in achieving improved results. Furthermore, this approach slightly improved the masking and data pretraining processes. Initially, we utilized RoBERTa-base and large models, following a similar approach to the pre-trained BERT models. Subsequently, we employed Twitter-RoBERTa-base for sentiment analysis, a model trained on approximately 58M tweets and fine-tuned for sentiment analysis using the TweetEval benchmark from \cite{TWEETEVAL2020}. Finally, we evaluated Persian RoBERTa, a model with a concept similar to ParsBERT but based on the RoBERTa architecture.

\medbreak
\subsubsection{Lite BERT for Self-supervised Learning of Language Representations (ALBERT)}

ALBERT, introduced by \cite{ALBERT2019}, introduced two significant innovations over BERT. First, it factorized embedding parameterization by using a smaller embedding size and projecting it to the transformer hidden size. Additionally, ALBERT shared all parameters between transformer layers. For our classification task, we used the Persian ALBERT v3.0 model, which was provided in ParsBERT.

\medbreak
\subsubsection{Distilled Version of BERT (DistilBERT)}

Distillation, as described by \cite{DISTILLATION2015}, involved training a small student model to mimic a larger teacher model as closely as possible. DistilBERT was introduced based on this concept \citep{DISTILBERT2019}. To integrate DistilBERT into the study, we used the Persian DistilBERT v3.0 model, as implemented by ParsBERT.

\medbreak
\subsubsection{Generalized Auto-regressive Pretraining for Language Understanding (XLNet)}

BERT had two main limitations. It distorted the input with masks and suffered from dissimilarity between pretraining and fine-tuning. Additionally, BERT ignored the dependency between masked positions. To address these issues, \cite{XLNET2019} introduced XLNET, employing a permutation language modeling idea. Furthermore, they employed techniques for masking and using the position of the prediction token. We used XLNet-base and XLNet-large pre-trained models to assess this architecture, evaluate the results, and compare them with other transformer-based models.

\medbreak
\subsubsection{Unsupervised Cross-lingual Representation Learning at Scale (XLM-R)}

In addition to monolingual models, we also fine-tuned and evaluated XLM-RoBERTa (XLM-R) from \cite{XLMR2019}, a transformer-based multilingual masked language model pre-trained on text in 100 languages. We utilized the XLM-RoBERTa-large model for this purpose.

\subsection{Emotion Analysis}

To analyze the emotion of vaccine-related tweets during the Covid-19 pandemic, we utilized a proprietary dataset that provided happiness and anger levels for a lexicon of 8,375 common Persian words frequently used on Twitter. Six individuals participated in the evaluation and labeling of this dataset. Each word in the lexicon was assigned two numbers between 1 and 9, representing the intensity of happiness and anger, with 5 indicating a neutral state and higher numbers denoting more extreme emotions. This method was similar to the Hedonometer approach proposed by \cite{HEDONOMETER2011} for measuring expressed happiness in other languages. We calculated average happiness and anger weights for each word in the dataset. Then, we fitted the inverse of the normal distribution function to assign weights to each number between 1 and 9. The purpose of using this function was to emphasize the impact of extremely emotional words.

Afterward, we employed the dataset to scale up the emotion analysis from individual words to texts. To evaluate the weighted average level of anger and happiness, we utilized an algorithm (H-AVG) based on Hedonometer's proposal, as follows:

$$
h_{\mathrm{avg}}(T)=\frac{\sum_{i=1}^{N} h_{\mathrm{avg}}\left (w_{i}\right) \times freq_{i}}{\sum_{i=1}^{N} freq_{i}}
$$

$$
a_{\mathrm{avg}}(T)=\frac{\sum_{i=1}^{N} a_{\mathrm{avg}}\left (w_{i}\right) \times freq_{i}}{\sum_{i=1}^{N} freq_{i}}
$$

where $freq_{i}$ is the frequency of the word $w_{i}$ ($i$th word) in text $T$, and $N$ is the number of words present in $T$.

Before calculating the averages, we excluded every word not found in the emotion dataset. Additionally, we eliminated all neutral words to emphasize the overall levels of happiness and anger in tweets. Next, we computed the average happiness and anger for each tweet while excluding any words not present in our original emotion dataset. To ensure more robust results, we opted to assign average happiness and anger scores for missing words, as shown below:

$$
h_{\mathrm{avg}}(w)=\frac{\sum_{i=1}^{M} h_{\mathrm{avg}}\left (T_{i}\right)}{\sum_{i=1}^{M} freq_{i}}
$$

$$
a_{\mathrm{avg}}(w)=\frac{\sum_{i=1}^{M} a_{\mathrm{avg}}\left (T_{i}\right)}{\sum_{i=1}^{M} freq_{i}}
$$

where $T_{i}$ is the $i$th text containing word $w$, and $M$ is the number of texts containing $w$.

Subsequently, we employed H-AVG once more to calculate the average happiness and anger per day during the Covid-19 pandemic and compared the results with Covid-19-related events in Iran. The results have been reported in Section \ref{sec:results}.

\subsection{Vaccine Themes}

Upon achieving an acceptable result (detailed in Section \ref{subsec:vaccine-classification}) for the classification of vaccine-related tweets, we extracted the main subjects from both vaccine opposition and support categories. Initially, we considered 500 randomly selected tweets from the combined groups. Subsequently, we employed a grounded theoretical approach and inductive analysis to manually identify the main themes. We analyzed and assigned relevant themes to each tweet and extracted essential keywords associated with each theme based on the tweet content. To focus solely on the primary subjects of each tweet, we considered at most three relevant themes for each. Then, for each theme identified in the vaccine opposition and support groups, we compiled a set of keywords that represented the subject's concept. These keywords were subsequently used to categorize the remaining tweets in each vaccine-related group. Our goal was to identify one or more themes for at least 85\% of the tweets (excluding neutral ones). To achieve this, we continued grouping tweets and adding extra categories. Once this objective was met, we had identified 15 distinct themes, each with its unique set of keywords, for the vaccine opposition group. Similarly, there were 16 distinct themes for the vaccine-supportive group of tweets. This meant that the core topic of 85\% of vaccine opposition and support tweets was identified using a total of 31 themes. The remaining 15\%, typically comprising short tweets (less than four words), had vague or unknown overall topics. Since we employed a keyword-based approach, the primary reason for not categorizing them into any pre-defined themes was the insufficient number of words. For example, \textit{How about Vaccination?} serves as a good example that did not convey any meaningful or subjective opinion on the subjects.


\subsection{User Interaction Analysis}

Evaluating user activities, especially for influencers (users with a high interaction rate), provided us with insights into user attitudes and changes in trends that were not perceivable through the assessment of tweets.

To evaluate user behavior toward Covid-19 vaccination, we initially categorized users on a monthly basis from February 2020 to December 2021 into four different groups: anti-vaccination, neutral, pro-vaccination, and mixed. If 60\% or more of a user's tweets about vaccination in a given month belonged to the vaccine opposition group, we categorized the user in the anti-vaccination group for that specific month. Similarly, we classified users into the pro-vaccination and neutral groups. In cases where a user could not be assigned to any specific group in a month based on these criteria and the mentioned threshold, we categorized them as mixed. The complete details of the method used for user classification are presented in Algorithm \ref{alg:cap}.

\begin{algorithm}
\caption{Single User Classification Algorithm}\label{alg:cap}
    \begin{algorithmic}

    \State $a \gets$ percentage of anti-vaccination tweets in a month
    \State $p \gets$ percentage of pro-vaccination tweets in a month
    \State $n \gets$ percentage of neutral tweets in a month
    
    \If{Any of the representative variables is greater than $60$}
        \State User is categorized accordingly
    \ElsIf{$a == 0$}    \Comment{40 $\leq$ p, n $\leq$ 60}
        \State User is classified as pro-vaccination
    \ElsIf{$p == 0$} \Comment{40 $\leq$ a, n $\leq$ 60}
        \State User is classified as anti-vaccination
    \Else
        \State User is classified as mixed
    \EndIf
\end{algorithmic}
\end{algorithm}

In the next step, we assessed the activities and interactions of influencers. We created a user interaction graph in which there was an edge between two users if one mentioned or replied to the other. The total number of connections for each user (node degree) was used as a metric to analyze a person's influence. By calculating the number of connections each user had per month, we identified the top 40 users with the highest degree for each of the 23 available months as the influencers (top 0.2 percent of each month's users). We then examined the distribution of influencers with respect to the four categories mentioned earlier. Finally, to gain an overview of the overall interactions and the impact of the vaccine program, we created two social networks using the user data: one before the start of public vaccination in Iran, which occurred on June 1, 2021, and the other after that date.

\section{Results}
\label{sec:results}

\subsection{Vaccine-related Tweets}

\begin{figure*}
    \centering
    \includegraphics[width=\textwidth]{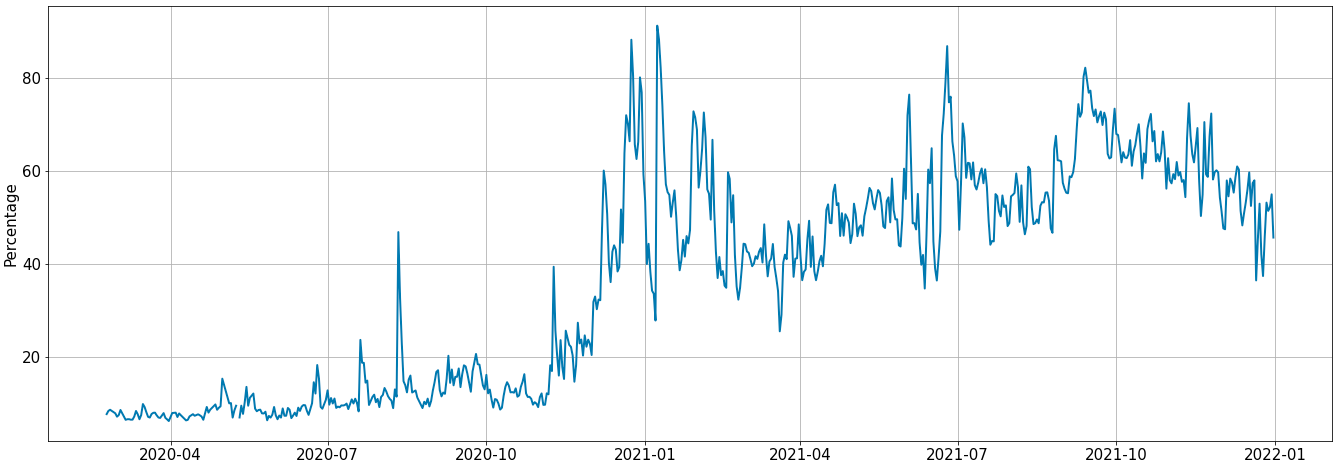}
    \caption{Relative Percentage of Vaccine Tweets Over Time} 
    \label{fig:vaccine-tweets-timeline} 
\end{figure*}

We collected 3,539,196 tweets relevant to Covid-19, and 1,037,440 of them were categorized as vaccine-related tweets (as shown in Figure \ref{fig:vaccine-tweets-timeline}) based on our hybrid approach described in Section \ref{sec:methods}.

From February 2020 to December 2021, an average of 37.65\% (median 42.09\%) of Covid-19 tweets per day were related to vaccination. To delve deeper into this analysis, we examined our data with respect to two significant dates: the introduction of Coronavirus vaccines on February 9, 2021, and the commencement of public vaccination in Iran on June 1, 2021.

According to our evaluations, as presented in Table \ref{tab:tweets-over-pandemic}, a greater proportion of the tweets after February 9, 2021, were related to vaccination compared to the earlier period of the Covid-19 pandemic. Similarly, following the start of public vaccination, the rate of vaccine-related tweets significantly exceeded that before this date. We observed that subsequent to the official introduction of vaccines and the commencement of public vaccination, there was a substantial increase in vaccine-related tweets, reflecting new topics related to vaccine matters, including side effects, effectiveness, and general opinions on vaccine uptake.

\begin{table}[ht] 
\caption{Vaccine-related Tweets over Covid-19 Pandemic} 
\centering
    \begin{tabular}{|c|c|}
\hline\hline
 Pandemic Period & \  Daily Avg. Vaccine-related Tweets \\ [0.25ex] 
\hline
Before 9 Feb. 2021 &  21.11\%\\
After 9 Feb. 2021 &  54.43\%\\
\hline
Before 1 Jun. 2021 &  27.29\%\\
After 1 Jun. 2021 &  58.71\%\\
\hline\hline  
\end{tabular} 
\label{tab:tweets-over-pandemic} 
\end{table}

\subsection{Vaccine-related Tweets Classification}
\label{subsec:vaccine-classification}

To classify our vaccine-related tweets into vaccine-opposition, neutral, and vaccine-support groups, we labeled 6000 tweets and subsequently randomly split our tagged data into train and validation sets. The training set comprised 5000 tweets, with the remainder allocated to the validation set. Additional details about the partition can be found in Table \ref{tab:train_eval_sets}.

\begin{table}[ht] 
\caption{Polarity Distribution in Training and Validation Sets} 
\centering
    \begin{tabular}{|c|c|c|c|}
\hline\hline
 Sets & \  Critical & \  Neutral & \  Supportive \\ [0.25ex] 
\hline
Training & 1467 & 2167 & 1366 \\
Validation & 268 & 444 & 288 \\
\hline\hline  
\end{tabular} 
\label{tab:train_eval_sets} 
\end{table}

Based on the dataset extension method described in Section \ref{sec:methods}, where we had four hyperparameters to tune, the best average result was achieved by the dataset with no character duplication removal, moderate odd pattern removal, no lemmatization, and no stopword elimination. We referred to this dataset as the final dataset. Subsequently, we proceeded to fine-tune our transformer-based models on the final dataset and compared their results to identify the best model for our classification task. Table \ref{tab:classification-results} provides more information about the top results of our classification. As seen, our fine-tuned Pars-BERT model outperformed all other approaches with an F1-Score of 62.03\%. Other models, such as BERT, RoBERTa, Twitter-RoBERTa, XLM-R, and XLNET, did not achieve an F1-Score exceeding 30\%.

\begin{table}[ht] 
\caption{Vaccine-related Tweets Classification Results} 
\centering
    \begin{tabular}{|c|c|c|c|}
    \hline\hline
     Models & \  F1-Score & \  Accuracy (O, N, S) \\ [0.25ex] 
    \hline
    Persian ALBERT & 39.9 & 77.22, 50.34, 4.24 \\
    Persian RoBERTa & 53.78 & 48.40, 61.78, 46.64 \\
    Persian DistilBERT & 58.45 & 51.25, 69.57, 49.12 \\
    \textbf{Pars-BERT} & \textbf{62.03} & \textbf{63.06, 60.81, 61.81} \\ 
    \hline\hline 
\end{tabular} 
\label{tab:classification-results} 
\end{table}

\subsection{Emotion Analysis}

\begin{figure}
    \centering
    \includegraphics[scale=0.3]{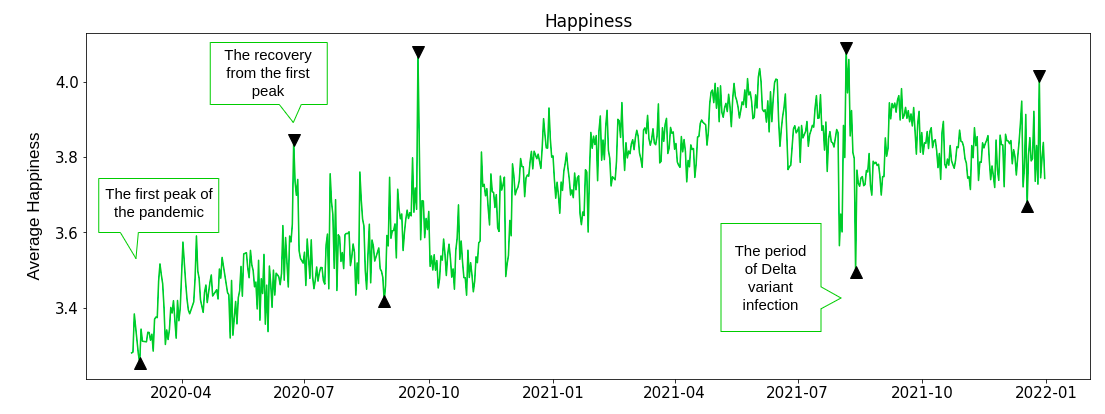}
    \caption{Happiness Trend of Vaccine-Related Tweets during Covid-19 Pandemic} 
    \label{fig:hap-emo} 
\end{figure}

\begin{figure}
    \centering
    \includegraphics[scale=0.3]{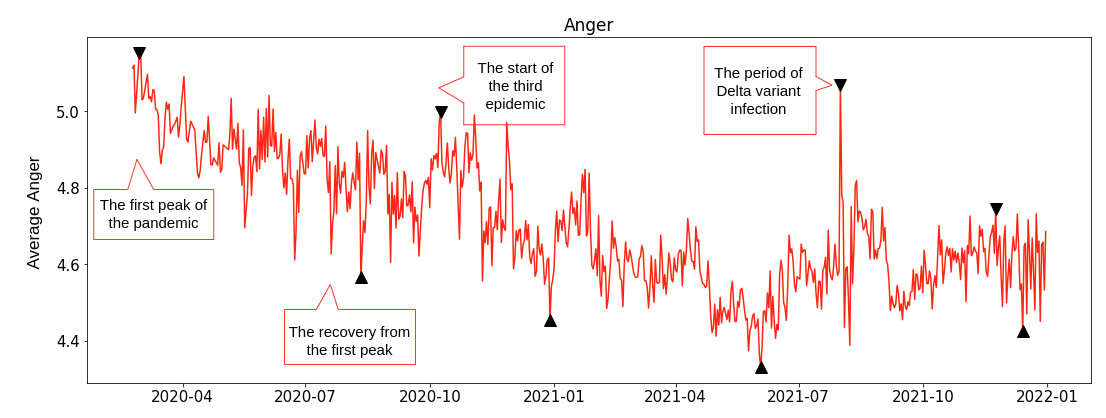}
    \caption{Anger Trend of Vaccine-Related Tweets during Covid-19 Pandemic} 
    \label{fig:ang-emo} 
\end{figure}

The results of tweet emotion detection are presented in Figures \ref{fig:hap-emo} and \ref{fig:ang-emo}. In these time series, several important dates (peaks and valleys) existed for each emotion type (indicated by black triangles). We cross-referenced these dates with the introduction of vaccines and two available time series, namely, the daily number of new cases and the number of deaths. We identified several interesting correlations, including:

\begin{itemize}
    \item March 2020 - April 2020: The first peak of the pandemic (First happiness valley and anger peak): During the initial worldwide shock of Covid-19 and with no vaccines or other treatments available, there was widespread public panic concerning the consequences of the Coronavirus.
    \item June 2020 - July 2020: Recovery from the first peak (First happiness peak and anger valley): Although no vaccination methodology had been discovered, the overall decline in the rate of Covid-19 infections led to the belief that the public was becoming less susceptible to the disease.
    \item October 2020 - November 2020: The start of the third epidemic (Second anger peak): The initiation of vaccination in other countries and reports on vaccine effectiveness, coupled with the critical situation and high infection rates in Iran, led to significant dissatisfaction and outrage among the public towards Covid-19.
    \item July 2021 - September 2021: The period of the Delta variant infection (Third happiness valley and anger peak): The Delta variant of the Coronavirus marked one of the most significant periods in terms of daily new cases and deaths. Although vaccination effectively controlled sad and angry sentiments, the last anger peak and happiness valley were more pronounced compared to other significant dates.
\end{itemize}

Furthermore, when examining the entire happiness time series, we observed an overall upward trend. This trend became evident when we compared the periods before and after the introduction of vaccines in February 2021. In contrast, the anger time series exhibited the opposite trend, showing a decline when comparing the averages before and after the introduction of vaccines. We assessed the correlation between the happiness and anger trends using Spearman's rho and Pearson's coefficients. The coefficients for both measures were -0.965 with a p-value $<$ 0.001, indicating a high negative correlation between these two trends.

As the results showed, vaccination had an impact on public happiness and anger towards Coronavirus. Due to the effectiveness of vaccines, people began to place more trust in vaccination as a remedy for Coronavirus, leading to a reduction in the number of sad or angry tweets related to Covid-19. Furthermore, we identified a high correlation between sadness and anger concerning Covid-19 vaccination. This finding serves as an example of how different emotions can be influenced in a similar manner by an external factor, such as a pandemic. It may also explain how the same negative or positive emotions can significantly reinforce each other when they align, as demonstrated in previous studies \citep{sahu2014depression, zhan2015distinctive}.



\subsection{Vaccine Themes}
Classified data were analyzed to extract themes for both vaccine-critical and supportive tweets. Following the extraction methods mentioned earlier, 219,646 tweets were labeled as having vaccine-opposition content. These tweets were distributed among 15 distinct categories, with an additional category named $other$. The same approach was applied to the vaccine-supportive tweets, resulting in 339,351 distinct tweets distributed across 16 different themes. Similar to the critical side, a category called $other$ was also included. Details of both themes are provided in Table \ref{tab:vaccine-themes}. Since we utilized a keyword-based approach, it was possible for a single tweet to belong to more than one category (in both themes). Therefore, the sum of frequencies for vaccine support and critical categories exceeded 100\%.

\begin{table}[ht] 
    \caption{Brief Description of Vaccine Themes}
    \centering
    \begin{tabular}{|p{0.23\linewidth}|p{0.37\linewidth}|p{0.12\linewidth}|p{0.14\linewidth}|}
    \hline\hline
     Theme Name & \  Description  & \  Critical & \  Supportive \\ [0.25ex] 
    \hline
    Side Effects & Mentions of health impacts caused by vaccines & 43,608 (19.85\%) & 46,551 (13.72\%) \\
    \hline
    Pharmaceuticals & Talks about vaccine names and companies making vaccines & 34,398 (15.66\%) & 47,506 (14.00\%) \\
    \hline
    Political / Governmental & Conversations on governmental actions towards mass vaccination & 94,748 (43.14\%) & 135,095 (39.81\%) \\
    \hline
    Vaccine Ingredients & Related to how vaccines are created and their materials & 10,293 (4.69\%) & 7,991 (2.35\%) \\
    \hline
    Research Trials & References to experiments and lab works & 26,394 (12.02\%) & 62,252 (18.42\%) \\
    \hline
    Religion & Topics on faith and religious practices & 9,793 (4.46\%) & 18,971 (5.59\%) \\
    \hline
    Ineffectiveness / Hesitancy & Conversations on low vaccine impression and incapability to fight Covid-19 & 50,639 (23.05\%) & - \\
    \hline
    Safety / Sufficiency & References to vaccine performance and ability & - & 88,627 (26.12\%) \\
    \hline
    Disease Prevalence & Mentions of virus mutations over time & 4,756 (2.17\%) & 20,843 (6.14\%) \\
    \hline
    Family & Expression of the concern for family members and relatives & 15,278 (6.96\%) & 28,253 (8.33\%) \\
    \hline
    Foreign Countries & Talks of pandemic state in other countries and imported vaccines & 93,478 (42.56\%) & 79,794 (23.51\%)  \\
    \hline
    Lockdown Denial & Related to ignoring the pandemic and worldwide crisis &  5,602 (2.55\%) & - \\
    \hline
    Pandemic Confirmation & Relevant to accepting the pandemic & - & 88,913 (26.20\%) \\
    \hline
    Mandatory vaccination & Criticism of forced vaccination and encouragements & 15,616 (7.11\%) & - \\
    \hline
    Influential Users & Mentions of influencers and their actions towards vaccination & 15,334 (6.98\%) & 31,970 (9.42\%) \\
    \hline
    Vaccine Alternatives & Other vaccine substitutes, their advantages and disadvantages & 4,914 (2.24\%) & 2,406 (0.71\%) \\
    \hline
    Medics and Hospitals & Relevant to doctors and other treatment staff & 37,053 (16.87\%) & 48,472 (14.28\%) \\
    \hline
    Hope / Envy & Expressions of impatience towards receiving vaccination & - & 37,142 (10.95\%) \\
    \hline
    Availability & Demanding public vaccination from authorities & - & 9,467 (2.79\%) \\
    \hline
    Others & Not categorized in any of themes & 31,953 (14.53\%) & 45,659 (13.45\%) \\
    
    \hline\hline  
    \end{tabular}
    \label{tab:vaccine-themes} 
\end{table}

\begin{figure}
    \centering
    \includegraphics[scale=0.51]{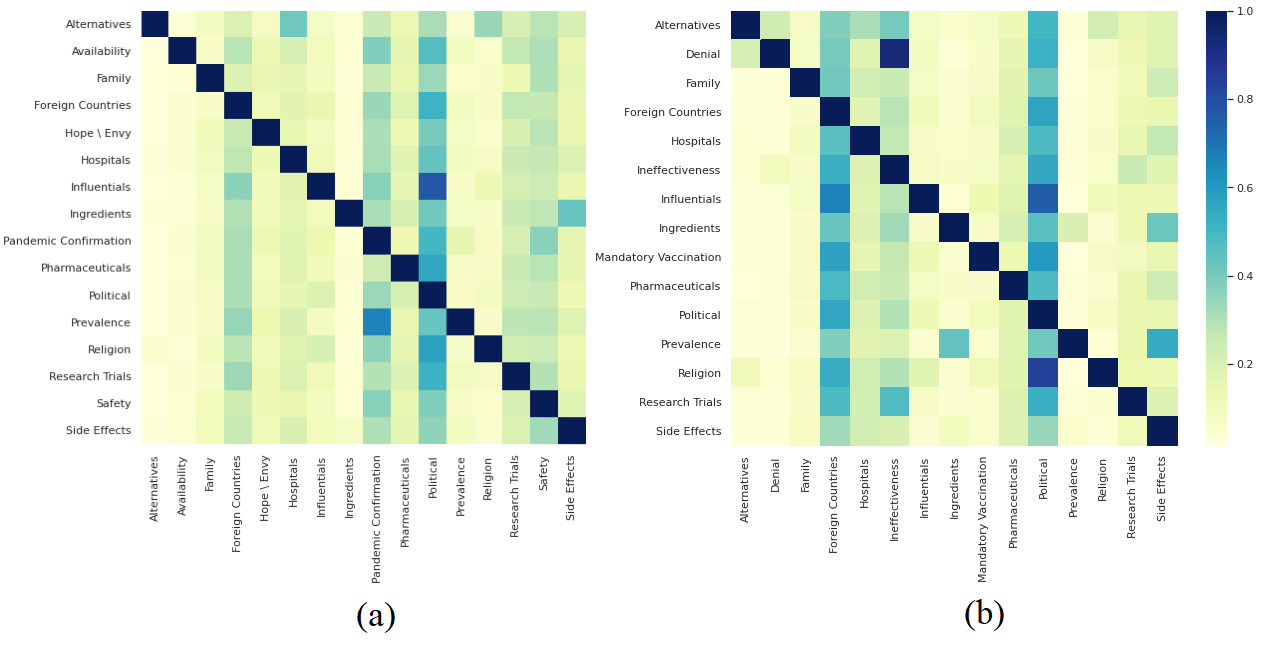}
    \caption{Support (a) and Opposition (b) Themes Correlation} 
    \label{fig:co-occ} 
\end{figure}

Figure \ref{fig:co-occ} illustrates the correlation of themes for both supportive and critical groups. The left correlation matrix belongs to supportive themes, and the right one represents the critical side. There are several strong relationships that are worth mentioning, which are as follows:
\begin{itemize}
    \item Influencers and Political (Both supportive and critical): Most of the tweets concerning influencers like actors and officials regard their reaction and decisions toward the Covid-19 based on the political situation in Iran.
    \item Prevalence and Pandemic Confirmation (Supportive): As the Covid-19 prevalence and mutations affect people increasingly, there is a higher rate of widespread pandemic acceptance and supportive opinions regarding taking vaccines.
    \item Ingredients and Side Effects (Supportive): It seems that talking about vaccine ingredients usually infers the matters impacting human health for a long or short time. That is why most of the contexts discussing ingredients also refer to the side effects in humans.
    \item Denial and Ineffectiveness (Critical): Ignoring the pandemic is alongside disregarding the Covid-19 crisis. On the one hand, people denying Coronavirus might also tend to deny vaccines and their effectiveness; on the other hand, they might consider both Covid-19 and vaccines a delusion.
    \item Religious and Political (Critical): Tweets containing spiritual concepts, Talk about Covid-19 from the religious viewpoint. According to the results, these tweets seem to relate the political decisions toward vaccination in Iran to religious instructions.
\end{itemize}

\subsection{User Interaction Analysis}

\begin{figure}
    \centering
    \includegraphics[scale=0.4]{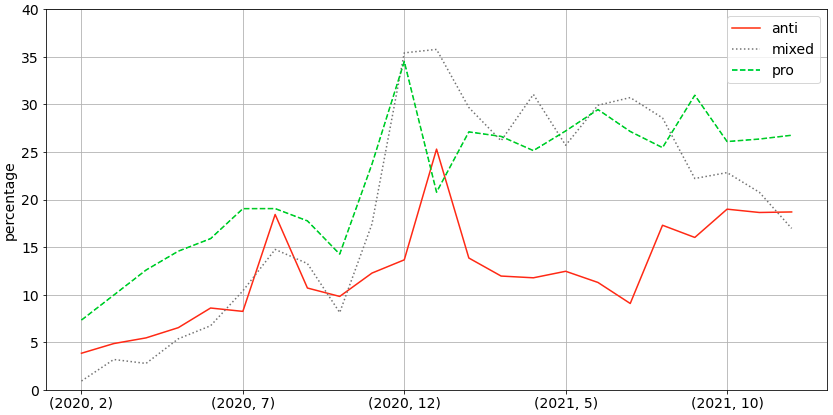}
    \caption{User Classes During Covid-19 Pandemic} 
    \label{fig:user-classification} 
\end{figure}

Assessing users' mindsets behind their tweets led us to categorize them into four different groups: anti-vaccination, pro-vaccination, neutral, and mixed. Figure \ref{fig:user-classification} presented the flow of changes in anti, pro, and mixed classes based on the relative percent of monthly coverage for each group during the Covid-19 pandemic.

As shown, between the introduction of Coronavirus vaccines in Iran and the beginning of public vaccination (from February to June 2021), the percentage of anti-vaccination users was 4.18\% lower compared to the months when vaccination was in progress. Similarly, the ratio of anti-vaccination users between February and June 2021 was 2.78\% lower than in the months before the vaccine introduction. On the other hand, an analysis of pro-vaccination users revealed that during the period from the vaccine introduction up to the end of 2021, the percentage of vaccination supporters increased by 9.67\% compared to the time before February 2021.

\begin{figure}
    \centering
    \includegraphics[scale=0.39]{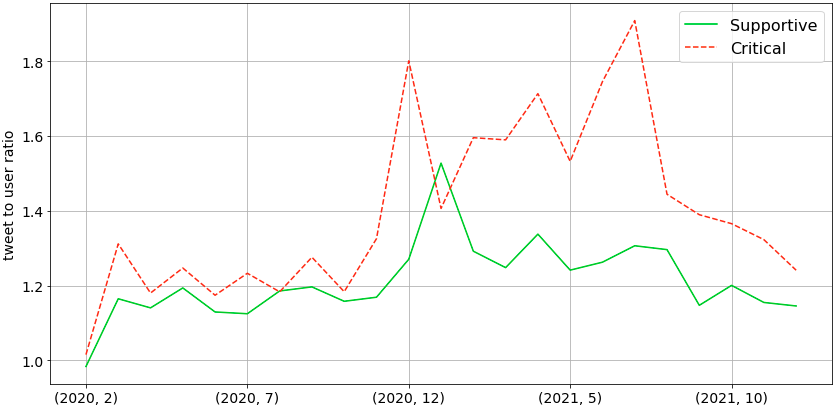}
    \caption{Tweet to User Ratio} 
    \label{fig:tweet-user-ratio} 
\end{figure}

By analyzing the results, we observed that vaccination and its results helped reduce criticism about vaccines. However, to evaluate the activity of each group, we calculated the ratio of the number of tweets to the number of users for both supportive and critical groups. According to Figure \ref{fig:tweet-user-ratio}, the average ratio for the critical group was 0.2 (15.7\%) higher than for the supportive group. This difference was even more significant during the period between the introduction of vaccines and the start of public vaccination. Based on the results, we can infer that during that period, people understood that vaccination was inevitable; hence, their opposition and hesitancy were even more pronounced. On the other side, those who agreed with vaccination represented their thoughts more widely than before.

After the initiation of public vaccination, there was a considerable fall in the rate of the critical group, indicating that the recovery results had convinced some critics to accept the efficiency of Covid-19 vaccines. However, inferring from the slight decline in the supportive group, it appears that the vaccination results were not as promising as expected.

\begin{figure}
    \centering
    \includegraphics[scale=0.25]{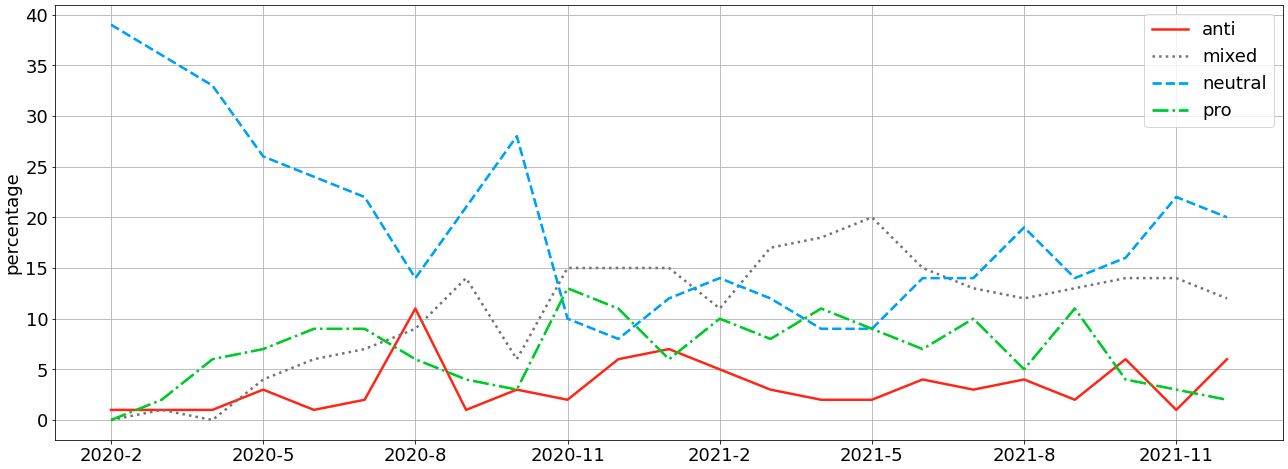}
    \caption{Top Influencers Per Month} 
    \label{fig:influencer-results} 
\end{figure}

In the next step, we evaluated influencers by considering user interactions, which included replies and mentions. As previously mentioned, the top 40 users with the highest rate of interaction for each month were labeled as influencers. Figure \ref{fig:influencer-results} shows the classification of such users during the pandemic. By looking at the number of influencers categorized as pro-vaccination and anti-vaccination, we discovered that vaccine-critical influencers made up 7.91\% of the whole influencer population before the introduction of Covid-19 vaccines in Iran. However, this share changed to 8.63\% afterward. As for the other side, vaccine supporters' coverage increased from 16.04\% to 18.18\%. From these observations, we can infer that the dissemination of vaccines resulted in more non-neutral tweets and conversations from influencers, as factors such as efficiency and side effects became much more apparent than before, and users became extra opinionated.

\begin{figure*}
    \centering
    \includegraphics[width=\textwidth]{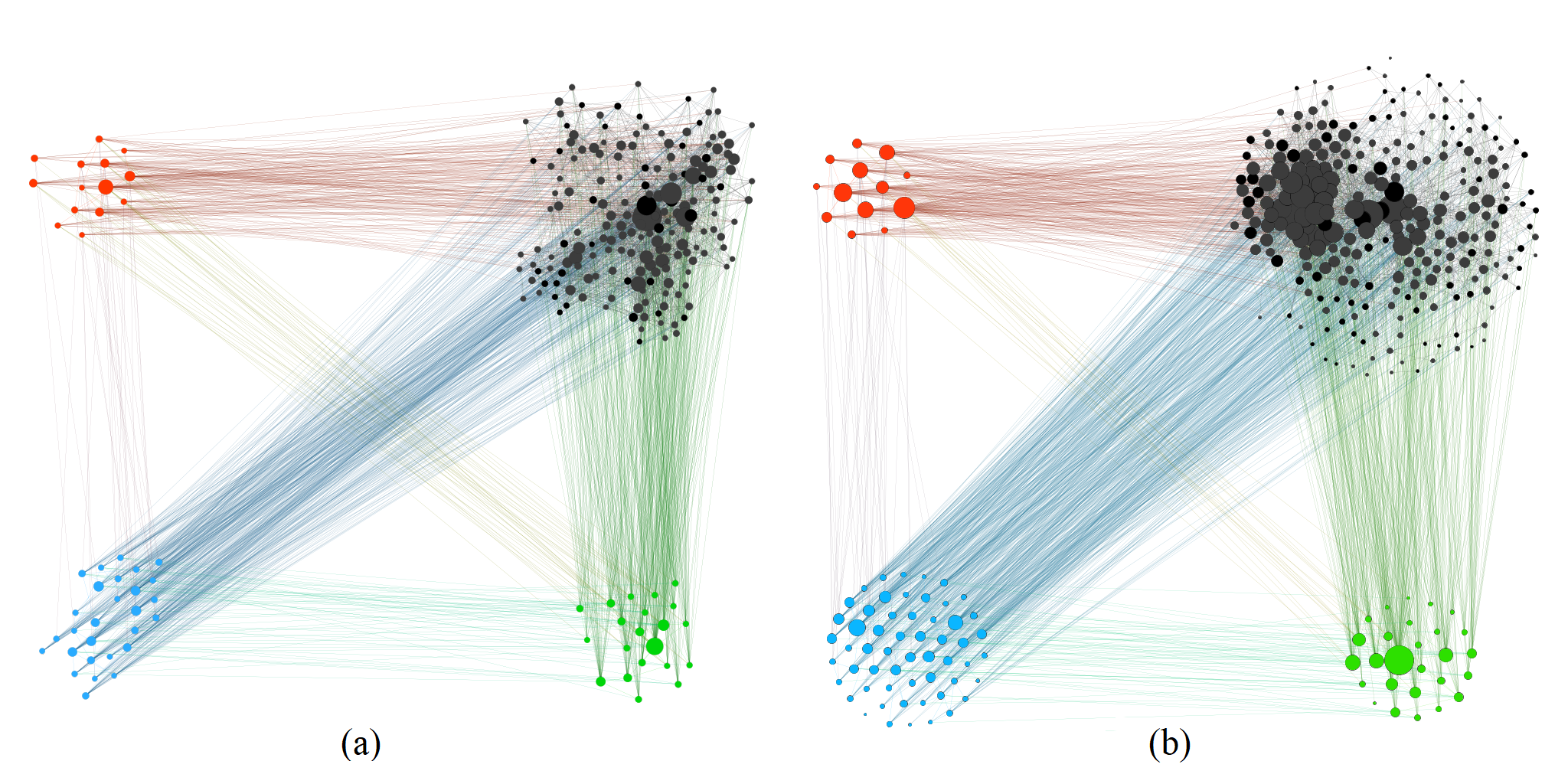}
    \caption{User Interactions before (a) and after (b) Public Vaccination in Iran. Red nodes showing anti-vaccinations, green ones for pro-vaccination, blue nodes representing neutral, mixed users are shown in gray, and black ones demonstrating unclassified users.} 
    \label{fig:before-after-vaccination}
\end{figure*}

In order to provide a summary of the overall interactions and the impact of the vaccination program, we created two networks. One for before ($BV$) and the other after ($AV$) the public vaccination (June 2021) in Iran, as represented in Figure \ref{fig:before-after-vaccination}. We excluded users who had fewer than 350 interactions in each of the two mentioned periods for these networks. Green nodes represent pro-vaccination, and red ones illustrate anti-vaccination. Neutral and mixed users appear as blue and gray nodes, respectively. Furthermore, some users were not found in our separately-gathered dataset of users, which might be non-Persian users mentioned or replied to by others; we specified them in black. Moreover, the number of connections is demonstrated with node diameter.

According to our evaluations, before June 2021, anti-vaccination users constituted 7.11\% of all users, while they formed only 4.73\% after that time. Likewise, pro-vaccination members accounted for 12.44\% before June 2021, whereas they made up for 9.82\% afterward. These two trends disagreed with what we had observed for the influencers, meaning that normal users from both sides of the argument became less fixated on their positions and, on average, decided to either post fewer amounts of content or take relatively neutral stands toward the vaccination.

Table \ref{tab:networks-measures} shows the overall statistics of both networks. Based on the in-degree and density measures, we observed that users tended to receive fewer mentions and replies after the public vaccination compared to the previous period. Similarly, the rate of contribution to vaccine-related tweets decreased. These results showed that after public vaccination and its observable effects, the level of Covid reactions in tweets decreased significantly. Nevertheless, the rate of top influencers (denoted with a large diameter) increased, especially for anti-vaccination and pro-vaccination users, showing that logical discussions among prominent members enhanced.

\begin{table}[ht] 
\caption{User Interaction Network Measures} 
\centering
    \begin{tabular}{|c|c|c|}
    \hline\hline
    Measures & \  BV & \   AV \\ [0.25ex] 
    \hline
    Average in-degree & 25.88 & 21.52 \\
    Clustering Coefficient & 0.237 & 0.230 \\
    Density & 0.107 & 0.062 \\
    Homophily & 0.083 & -0.097 \\
    Average Path Length & 2.07 & 2.41 \\
    \hline\hline 
\end{tabular} 
\label{tab:networks-measures} 
\end{table}

Furthermore, consistent with the attachment of similar nodes demonstrated with the homophily measure, we observed that before vaccination, the argument between those with similar thoughts, affected by influencers such as news accounts, was bold. On the other side, as vaccination brought about healing outcomes and side-effects, the controversy among different groups with different viewpoints was raised after vaccination.

\section{Conclusion}
\label{sec:conclusion}

In this study, we used a keyword-based method to extract Covid-19-related tweets and performed topic modeling to specify the main subjects discussed around the Covid-19 matter. By utilizing the topic modeling results combined with a keyword-based search, we obtained vaccine-related tweets during the Coronavirus pandemic up to the end of 2021 in Iran. Subsequently, we classified vaccine-related tweets into vaccine-critical, neutral, and vaccine-supportive groups and extracted the main themes discussed around Covid-19 vaccination.

Moreover, we conducted a happiness and anger analysis to further evaluate public opinion toward vaccination. Afterwards, we carried out a range of analyses to assess how users reacted to the evolution of vaccines for Covid-19. The results demonstrate the immense potential of online platforms to provide insight into people's reactions to crises and how their behavior evolves. Although utilizing data from such platforms to understand the public response to Covid-19 has been explored to a certain degree, this study is among the first to address the issue in the Persian language. Future work can focus on more comprehensive analyses of network properties and structures, such as community detection, to gain a richer understanding of influential users and their connections. Furthermore, we did not segregate real accounts from fake users and bots. An accurate methodology to exclude bots from the user base would be beneficial for more robust insights into user behavior. Another important topic related to bots is their influence in steering society's way of thinking about vaccination and social matters in general. Studying their presence, attributes, or features that separate them from normal users, and the content they're spreading can be explored in the future to provide more cohesive and reliable information for people seeking information.

\section{Discussion}
\label{sec:discussion}

Our analysis of public opinion surrounding Covid-19 vaccination in Iran offers valuable insights while acknowledging certain limitations. Primarily, our data source is Twitter, which may not fully represent the entire population's perspectives. Additionally, we didn't differentiate between real user accounts and bots, which could impact data authenticity. Addressing these limitations in future research through more extensive data sources and bot detection methods will strengthen the reliability of our findings.

From a practical standpoint, our study carries implications for policymakers, healthcare practitioners, and communication experts. A deeper understanding of public sentiment towards vaccination can inform more effective vaccination campaigns, misinformation countermeasures, and public health interventions. Our research also contributes to the existing body of knowledge by examining the Persian language context, an area with relatively limited prior exploration. Leveraging topic modeling and emotion analysis, we uncovered evolving themes and emotional dynamics surrounding vaccination. This insight can help stakeholders tailor their messaging to better engage and educate the public about the importance of Covid-19 vaccination.










\bibliography{snbibliography}

\begin{thebibliography}{42}
\providecommand{\natexlab}[1]{#1}
\providecommand{\url}[1]{{#1}}
\providecommand{\urlprefix}{URL }
\providecommand{\doi}[1]{\url{https://doi.org/#1}}
\providecommand{\eprint}[2][]{\url{#2}}
 \bibcommenthead

\bibitem[{Barbieri et~al(2020)Barbieri, Camacho{-}Collados, Neves, and Anke}]{TWEETEVAL2020}
Barbieri F, Camacho{-}Collados J, Neves L, et~al (2020) Tweeteval: Unified benchmark and comparative evaluation for tweet classification. CoRR abs/2010.12421. \urlprefix\url{https://arxiv.org/abs/2010.12421}, {\href{https://arxiv.org/abs/2010.12421}{{https://arxiv.org/abs/2010.12421}}}

\bibitem[{Blei et~al(2003)Blei, Ng, and Jordan}]{LDA2003993}
Blei DM, Ng AY, Jordan MI (2003) Latent dirichlet allocation. J Mach Learn Res 3:993--1022. \doi{http://dx.doi.org/10.1162/jmlr.2003.3.4-5.993}, \urlprefix\url{http://portal.acm.org/citation.cfm?id=944937}

\bibitem[{Bonnevie et~al(2020)Bonnevie, Goldbarg, Gallegos-Jeffrey, Rosenberg, Wartella, and Smyser}]{BONNEVIE2020S326}
Bonnevie E, Goldbarg J, Gallegos-Jeffrey AK, et~al (2020) Content themes and influential voices within vaccine opposition on twitter, 2019. American Journal of Public Health 110(S3):S326--S330. \doi{10.2105/AJPH.2020.305901}, \urlprefix\url{https://doi.org/10.2105/AJPH.2020.305901}, pMID: 33001733, {\href{https://arxiv.org/abs/https://doi.org/10.2105/AJPH.2020.305901}{{https://arxiv.org/abs/https://doi.org/10.2105/AJPH.2020.305901}}}

\bibitem[{Bonnevie et~al(2021{\natexlab{a}})Bonnevie, Gallegos-Jeffrey, Goldbarg, Byrd, and Smyser}]{Bonnevie2021}
Bonnevie E, Gallegos-Jeffrey A, Goldbarg J, et~al (2021{\natexlab{a}}) Quantifying the rise of vaccine opposition on twitter during the covid-19 pandemic. Journal of Communication in Healthcare 14(1):12--19. \doi{10.1080/17538068.2020.1858222}

\bibitem[{Bonnevie et~al(2021{\natexlab{b}})Bonnevie, Gallegos-Jeffrey, Goldbarg, Byrd, and Smyser}]{BONNEVIE202112}
Bonnevie E, Gallegos-Jeffrey A, Goldbarg J, et~al (2021{\natexlab{b}}) Quantifying the rise of vaccine opposition on twitter during the covid-19 pandemic. Journal of Communication in Healthcare 14(1):12--19. \doi{10.1080/17538068.2020.1858222}, \urlprefix\url{https://doi.org/10.1080/17538068.2020.1858222}, {\href{https://arxiv.org/abs/https://doi.org/10.1080/17538068.2020.1858222}{{https://arxiv.org/abs/https://doi.org/10.1080/17538068.2020.1858222}}}

\bibitem[{Cascini et~al(2022)Cascini, Pantovic, Al-Ajlouni, Failla, Puleo, Melnyk, Lontano, and Ricciardi}]{cascini2022social}
Cascini F, Pantovic A, Al-Ajlouni YA, et~al (2022) Social media and attitudes towards a covid-19 vaccination: A systematic review of the literature. EClinicalMedicine

\bibitem[{Chang et~al(2009)Chang, Gerrish, Wang, Boyd-graber, and Blei}]{NIPS2009_f92586a2}
Chang J, Gerrish S, Wang C, et~al (2009) Reading tea leaves: How humans interpret topic models. In: Bengio Y, Schuurmans D, Lafferty J, et~al (eds) Advances in Neural Information Processing Systems, vol~22. Curran Associates, Inc., \urlprefix\url{https://proceedings.neurips.cc/paper/2009/file/f92586a25bb3145facd64ab20fd554ff-Paper.pdf}

\bibitem[{Chopra et~al(2021)Chopra, Vashishtha, Pal, Ashima, Tyagi, and Sethi}]{CHOPRA2021}
Chopra H, Vashishtha A, Pal R, et~al (2021) Mining trends of {COVID-19} vaccine beliefs on twitter with lexical embeddings. CoRR abs/2104.01131. \urlprefix\url{https://arxiv.org/abs/2104.01131}, {\href{https://arxiv.org/abs/2104.01131}{{https://arxiv.org/abs/2104.01131}}}

\bibitem[{Conneau et~al(2019)Conneau, Khandelwal, Goyal, Chaudhary, Wenzek, Guzm{\'{a}}n, Grave, Ott, Zettlemoyer, and Stoyanov}]{XLMR2019}
Conneau A, Khandelwal K, Goyal N, et~al (2019) Unsupervised cross-lingual representation learning at scale. CoRR abs/1911.02116. \urlprefix\url{http://arxiv.org/abs/1911.02116}, {\href{https://arxiv.org/abs/1911.02116}{{https://arxiv.org/abs/1911.02116}}}

\bibitem[{Devlin et~al(2018)Devlin, Chang, Lee, and Toutanova}]{BERT2018}
Devlin J, Chang M, Lee K, et~al (2018) {BERT:} pre-training of deep bidirectional transformers for language understanding. CoRR abs/1810.04805. \urlprefix\url{http://arxiv.org/abs/1810.04805}, {\href{https://arxiv.org/abs/1810.04805}{{https://arxiv.org/abs/1810.04805}}}

\bibitem[{Dodds et~al(2011)Dodds, Harris, Kloumann, Bliss, and Danforth}]{HEDONOMETER2011}
Dodds PS, Harris KD, Kloumann IM, et~al (2011) Temporal patterns of happiness and information in a global social network: Hedonometrics and twitter. PLOS ONE 6(12):1--1. \doi{10.1371/journal.pone.0026752}, \urlprefix\url{https://doi.org/10.1371/journal.pone.0026752}

\bibitem[{Durmaz and Hengirmen(2022)}]{durmaz2022}
Durmaz N, Hengirmen E (2022) The dramatic increase in anti-vaccine discourses during the covid-19 pandemic: a social network analysis of twitter. Human Vaccines \& Immunotherapeutics 18(1):2025,008. \doi{10.1080/21645515.2021.2025008}, pMID: 35113767

\bibitem[{Farahani et~al(2021)Farahani, Gharachorloo, Farahani, and Manthouri}]{ParsBERT}
Farahani M, Gharachorloo M, Farahani M, et~al (2021) Parsbert: Transformer-based model for persian language understanding. Neural Processing Letters \doi{10.1007/s11063-021-10528-4}

\bibitem[{Filter(2022)}]{site:CLEAN_TEXT}
Filter J (2022) {Functions to preprocess and normalize text}. \url{https://pypi.org/project/clean-text/}, [Online; accessed 21-April-2022]

\bibitem[{HAZM(2018)}]{site:HAZM}
HAZM (2018) {Python library for digesting Persian text}. \url{https://github.com/sobhe/hazm}, [Online; accessed 21-April-2022]

\bibitem[{Hinton et~al(2015)Hinton, Vinyals, and Dean}]{DISTILLATION2015}
Hinton G, Vinyals O, Dean J (2015) Distilling the knowledge in a neural network. \doi{10.48550/ARXIV.1503.02531}, \urlprefix\url{https://arxiv.org/abs/1503.02531}

\bibitem[{Hosseini et~al(2020)Hosseini, Hosseini, and Broniatowski}]{HOSSEINI2020}
Hosseini P, Hosseini P, Broniatowski DA (2020) Content analysis of persian/farsi tweets during {COVID-19} pandemic in iran using {NLP}. CoRR abs/2005.08400. \urlprefix\url{https://arxiv.org/abs/2005.08400}, {\href{https://arxiv.org/abs/2005.08400}{{https://arxiv.org/abs/2005.08400}}}

\bibitem[{Hutto and Gilbert(2014)}]{Hutto_Gilbert_2014}
Hutto C, Gilbert E (2014) Vader: A parsimonious rule-based model for sentiment analysis of social media text. Proceedings of the International AAAI Conference on Web and Social Media 8(1):216--225. \urlprefix\url{https://ojs.aaai.org/index.php/ICWSM/article/view/14550}

\bibitem[{Khan(2014)}]{grounded2014}
Khan S (2014) Qualitative research method: Grounded theory. International Journal of Business and Management 9. \doi{10.5539/ijbm.v9n11p224}

\bibitem[{Kharazi(2021)}]{site:PERSIAN_STOPWORDS}
Kharazi V (2021) {Persian Stop Words List}. \url{https://github.com/kharazi/persian-stopwords}, [Online; accessed 21-April-2022]

\bibitem[{Kwok et~al(2021)Kwok, Vadde, and Wang}]{kwok2021tweet}
Kwok SWH, Vadde SK, Wang G (2021) Tweet topics and sentiments relating to covid-19 vaccination among australian twitter users: machine learning analysis. Journal of medical Internet research 23(5):e26,953

\bibitem[{Lan et~al(2019)Lan, Chen, Goodman, Gimpel, Sharma, and Soricut}]{ALBERT2019}
Lan Z, Chen M, Goodman S, et~al (2019) {ALBERT:} {A} lite {BERT} for self-supervised learning of language representations. CoRR abs/1909.11942. \urlprefix\url{http://arxiv.org/abs/1909.11942}, {\href{https://arxiv.org/abs/1909.11942}{{https://arxiv.org/abs/1909.11942}}}

\bibitem[{Le et~al(2020)Le, Andreadakis, Kumar, Rom{\'{a}}n, Tollefsen, Saville, and Mayhew}]{ThanhLe2020}
Le TT, Andreadakis Z, Kumar A, et~al (2020) The {COVID}-19 vaccine development landscape. Nature Reviews Drug Discovery 19(5):305--306. \doi{10.1038/d41573-020-00073-5}, \urlprefix\url{https://doi.org/10.1038/d41573-020-00073-5}

\bibitem[{Liu et~al(2019)Liu, Ott, Goyal, Du, Joshi, Chen, Levy, Lewis, Zettlemoyer, and Stoyanov}]{ROBERTA2019}
Liu Y, Ott M, Goyal N, et~al (2019) Roberta: {A} robustly optimized {BERT} pretraining approach. CoRR abs/1907.11692. \urlprefix\url{http://arxiv.org/abs/1907.11692}, {\href{https://arxiv.org/abs/1907.11692}{{https://arxiv.org/abs/1907.11692}}}

\bibitem[{Lyu et~al(2022)Lyu, Wang, Wu, Duong, Zhang, Dye, and Luo}]{lyu2022social}
Lyu H, Wang J, Wu W, et~al (2022) Social media study of public opinions on potential covid-19 vaccines: informing dissent, disparities, and dissemination. Intelligent medicine 2(1):1--12

\bibitem[{Lyu et~al(2021)Lyu, Han, and Luli}]{LYE2021E24435}
Lyu JC, Han EL, Luli GK (2021) Covid-19 vaccine--related discussion on twitter: Topic modeling and sentiment analysis. J Med Internet Res 23(6):e24,435. \doi{10.2196/24435}, \urlprefix\url{https://www.jmir.org/2021/6/e24435}

\bibitem[{Newman et~al(2010)Newman, Lau, Grieser, and Baldwin}]{CV2010100}
Newman D, Lau J, Grieser K, et~al (2010) Automatic evaluation of topic coherence. pp 100--108

\bibitem[{Nezhad and Deihimi(2022)}]{NEZHAD2022}
Nezhad ZB, Deihimi MA (2022) Analyzing iranian opinions toward covid-19 vaccination. IJID Regions \doi{https://doi.org/10.1016/j.ijregi.2021.12.011}, \urlprefix\url{https://www.sciencedirect.com/science/article/pii/S2772707622000030}

\bibitem[{Organization(2021)}]{site:who2021}
Organization WH (2021) {Listings of WHO’s response to COVID-19}. \url{https://www.who.int/news/item/29-06-2020-covidtimeline}, [Online; accessed 10-April-2022]

\bibitem[{Sahu et~al(2014)Sahu, Gupta, and Chatterjee}]{sahu2014depression}
Sahu A, Gupta P, Chatterjee B (2014) Depression is more than just sadness: a case of excessive anger and its management in depression. Indian journal of psychological medicine 36(1):77--79

\bibitem[{Sanh et~al(2019)Sanh, Debut, Chaumond, and Wolf}]{DISTILBERT2019}
Sanh V, Debut L, Chaumond J, et~al (2019) Distilbert, a distilled version of {BERT:} smaller, faster, cheaper and lighter. CoRR abs/1910.01108. \urlprefix\url{http://arxiv.org/abs/1910.01108}, {\href{https://arxiv.org/abs/1910.01108}{{https://arxiv.org/abs/1910.01108}}}

\bibitem[{Shokrollahi et~al(2021)Shokrollahi, Hashemi, and Dehghani}]{SHOKROLLAHI2021}
Shokrollahi O, Hashemi N, Dehghani M (2021) Discourse analysis of covid-19 in persian twitter social networks using graph mining and natural language processing. CoRR abs/2109.00298. \urlprefix\url{https://arxiv.org/abs/2109.00298}, {\href{https://arxiv.org/abs/2109.00298}{{https://arxiv.org/abs/2109.00298}}}

\bibitem[{Smedt and Daelemans(2012)}]{JMLR:v13}
Smedt TD, Daelemans W (2012) Pattern for python. Journal of Machine Learning Research 13(66):2063--2067. \urlprefix\url{http://jmlr.org/papers/v13/desmedt12a.html}

\bibitem[{Thelwall et~al(2021)Thelwall, Kousha, and Thelwall}]{Thelwall_Kousha_Thelwall_2021}
Thelwall M, Kousha K, Thelwall S (2021) Covid-19 vaccine hesitancy on english-language twitter. Profesional de la Información 30(2). \doi{10.3145/epi.2021.mar.12}, \urlprefix\url{https://revista.profesionaldelainformacion.com/index.php/EPI/article/view/86322}

\bibitem[{Troiano and Nardi(2021)}]{TROIANO2021245}
Troiano G, Nardi A (2021) Vaccine hesitancy in the era of covid-19. Public Health 194:245--251. \doi{https://doi.org/10.1016/j.puhe.2021.02.025}, \urlprefix\url{https://www.sciencedirect.com/science/article/pii/S0033350621000834}

\bibitem[{Villavicencio et~al(2021)Villavicencio, Macrohon, Inbaraj, Jeng, and Hsieh}]{Philippines2021}
Villavicencio C, Macrohon JJ, Inbaraj XA, et~al (2021) Twitter sentiment analysis towards covid-19 vaccines in the philippines using naïve bayes. Information 12(5). \doi{10.3390/info12050204}, \urlprefix\url{https://www.mdpi.com/2078-2489/12/5/204}

\bibitem[{Wicke and Bolognesi(2021)}]{WICKE2021}
Wicke P, Bolognesi MM (2021) Covid-19 discourse on twitter: How the topics, sentiments, subjectivity, and figurative frames changed over time. Frontiers in Communication 6. \doi{10.3389/fcomm.2021.651997}, \urlprefix\url{https://www.frontiersin.org/article/10.3389/fcomm.2021.651997}

\bibitem[{Yang et~al(2019)Yang, Dai, Yang, Carbonell, Salakhutdinov, and Le}]{XLNET2019}
Yang Z, Dai Z, Yang Y, et~al (2019) Xlnet: Generalized autoregressive pretraining for language understanding. CoRR abs/1906.08237. \urlprefix\url{http://arxiv.org/abs/1906.08237}, {\href{https://arxiv.org/abs/1906.08237}{{https://arxiv.org/abs/1906.08237}}}

\bibitem[{Yin and Wang(2014)}]{GSDMM2014233}
Yin J, Wang J (2014) A dirichlet multinomial mixture model-based approach for short text clustering. Association for Computing Machinery, New York, NY, USA, KDD '14, p 233–242, \doi{10.1145/2623330.2623715}, \urlprefix\url{https://doi.org/10.1145/2623330.2623715}

\bibitem[{Yousefinaghani et~al(2021)Yousefinaghani, Dara, Mubareka, Papadopoulos, and Sharif}]{YOUSEFINAGHANI2021256}
Yousefinaghani S, Dara R, Mubareka S, et~al (2021) An analysis of covid-19 vaccine sentiments and opinions on twitter. International Journal of Infectious Diseases 108:256--262. \doi{https://doi.org/10.1016/j.ijid.2021.05.059}, \urlprefix\url{https://www.sciencedirect.com/science/article/pii/S1201971221004628}

\bibitem[{Zacharias(2020)}]{site:TWINT}
Zacharias C (2020) {Twitter Intelligence Tool}. \url{https://pypi.org/project/twint/}, [Online; accessed 19-April-2022]

\bibitem[{Zhan et~al(2015)Zhan, Ren, Fan, and Luo}]{zhan2015distinctive}
Zhan J, Ren J, Fan J, et~al (2015) Distinctive effects of fear and sadness induction on anger and aggressive behavior. Frontiers in Psychology 6:725

\end{thebibliography}

\end{document}